\def\year{2019}\relax
\documentclass[letterpaper]{article} 
\usepackage{aaai19}  
\usepackage{times}  
\usepackage{helvet}  
\usepackage{courier}  
\usepackage{url}  
\usepackage{graphicx}  

\usepackage{color}
\usepackage{subfigure}
\usepackage{amsmath}
\usepackage{multirow}
\usepackage{mathrsfs}

\begin{document}
%
\title{StNet: Local and Global Spatial-Temporal Modeling for Action Recognition}
\author{ \begin{tabular}{cccc} Dongliang He \textsuperscript{1} & Zhichao Zhou \textsuperscript{1} & Chuang Gan \textsuperscript{2} &  Fu Li \textsuperscript{1} \\
                                         Xiao Liu \textsuperscript{1} & Yandong Li \textsuperscript{3} &  Limin Wang \textsuperscript{4} &  Shilei Wen \textsuperscript{1}\end{tabular} \\ \\
          Department of Computer Vision Technology (VIS), Baidu Inc. \textsuperscript{1} \\
           MIT-IBM Watson AI Lab \textsuperscript{2}, University of Central Florida \textsuperscript{3}\\
           State Key Lab for Novel Software Technology, Nanjing University, China \textsuperscript{4} \\
          \{hedongliang01, zhouzhichao01, lifu, liuxiao12, wenshilei\}@baidu.com \\
           ganchuang1990@gmail.com,~~lyndon.leeseu@outlook.com,~~lmwang.nju@gmail.com}

\maketitle
\begin{abstract}
Despite the success of deep learning for static image understanding, it remains unclear what are the most effective network architectures for the spatial-temporal modeling in videos. In this paper, in contrast to the existing CNN+RNN or pure 3D convolution based approaches, we explore a novel spatial temporal network (StNet) architecture for both local and global spatial-temporal modeling in videos. Particularly, StNet stacks $N$ successive video frames into a \emph{super-image} which has $3N$ channels and applies 2D convolution on super-images to capture local spatial-temporal relationship. To model global spatial-temporal relationship, we apply temporal convolution on the local spatial-temporal feature maps. Specifically, a novel temporal Xception block is proposed in StNet. It employs a separate channel-wise and temporal-wise convolution over the feature sequence of video. Extensive experiments on the Kinetics dataset demonstrate that our framework outperforms several state-of-the-art approaches in action recognition and can strike a satisfying trade-off between recognition accuracy and model complexity. We further demonstrate the generalization performance of the leaned video representations on the UCF101 dataset.
\end{abstract}

\section{Introduction}
\begin{figure}[!ht]
\begin{center}
\includegraphics[width=0.9\columnwidth]{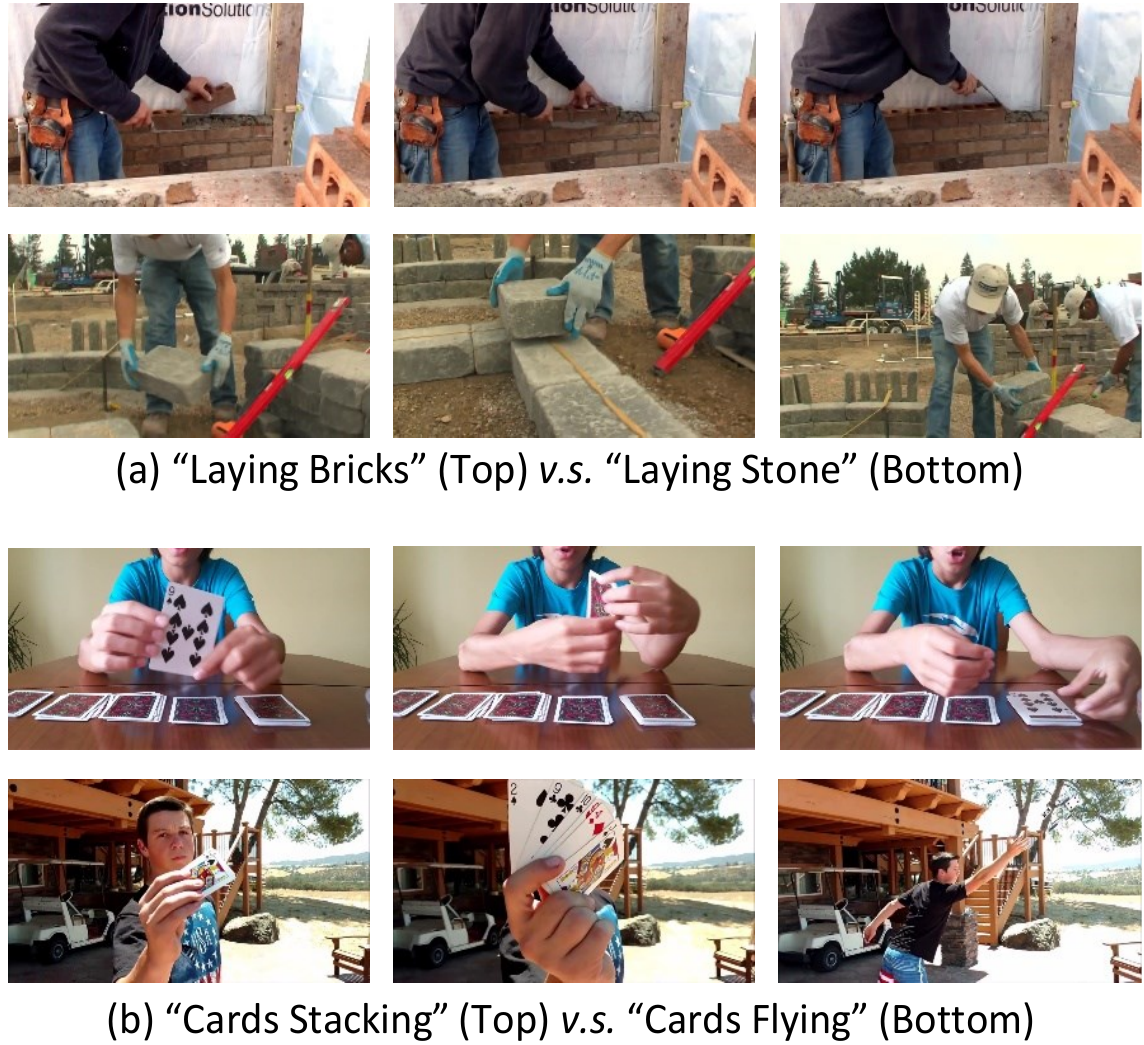}
\end{center}
   \caption{Local information is sufficient to distinguish ``Laying Bricks'' and ``Laying Stones" while global spatial-temporal clue is necessary to tell ``Cards Stacking'' and ``Cards Flying''. }
\label{fig:illustration}
\end{figure}

Action recognition in videos has received significant research attention in the computer vision and machine learning community \cite{karpathy2014,idt,two-stream,videodarwin,tsn,p3d,i3d,shuttleNet}.
The increasing ubiquity of recording devices has created videos far surpassing what we can manually handle. It is therefore desirable to develop automatic video understanding algorithms for various applications, such as video recommendation, human behavior analysis, video surveillance and so on. Both local and global information is important for this task, as shown in Fig.\ref{fig:illustration}. For example, to recognize ``Laying Bricks'' and ``Laying Stones'', local spatial information is critical to distinguish bricks and stones; and to classify ``Cards Stacking'' and ``Cards Flying'', global spatial-temporal clues are the key evidence.

Motivated by the promising results of deep networks \cite{bn,resnet,inceptionresnet} on image understanding tasks, deep learning is applied to the problem of video understanding. Two major research directions are explored specifically for action recognition, i.e., employing CNN+RNN architectures for video sequence modeling \cite{LRCN,yue2015beyond} and purely deploying ConvNet-based architectures for video recognition \cite{two-stream,two-stream-stresnet,T-Resnet,tsn,c3d,i3d,p3d}.

Although considerable progress has been made, current action recognition approaches still fall behind human performance in terms of action recognition accuracy. The main challenge lies in extracting discriminative spatial-temporal representations from videos. For the CNN+RNN solution, the feed-forward CNN part is used for spatial modeling, while the temporal modeling part, i.e., LSTM \cite{lstm} or GRU \cite{gru}, makes end-to-end optimization very difficult due to its recurrent architecture. Nevertheless, separately training CNN and RNN parts is not optimal for integrated spatial-temporal representation learning.

ConvNets for action recognition can be generally categorized into 2D ConvNet and 3D ConvNet. 2D convolution architectures \cite{two-stream,tsn} extract appearance features from sampled RGB frames, which only exploit local spatial information rather than local spatial-temporal information. As for the temporal dynamics, they simply fuse the classification scores obtained from several snippets. Although averaging classification scores of several snippets is straightforward and efficient, it is probably less effective for capturing spatio-temporal information. C3D \cite{c3d} and I3D \cite{i3d} are typical 3D convolution based methods which simultaneously model spatial and temporal structure and achieve satisfying recognition performance. As we know, compared to deeper network, shallow network exhibits inferior capacity on learning representation from large scale datasets. When it comes to large scale human action recognition, on one hand, inflating shallow 2D ConvNets to their 3D counterparts may be not capable enough of generating discriminative video descriptors; on the other hand, 3D versions of deep 2D ConvNets will result in too big model as well as too heavy computation cost both in training and inference phases.

Given the aforementioned concerns, we propose our novel spatial-temporal network (StNet) to tackle the large scale action recognition problem. First, we consider local spatial-temporal relationship by applying 2D convolution on 3N-channel \emph{super-image}, which is composed of N successive video frames. Thus local spatial-temporal information can be more efficiently encoded compared to 3D convolution on N images.
Second, StNet inserts temporal convolutions upon feature maps of super-images to capture temporal relationship among them. Local spatial-temporal modeling followed by temporal convolution can progressively builds global spatial-temporal relationship and is lightweight and computational friendly.
Third, in StNet, the temporal dynamics are further encoded with our proposed temporal Xception block (TXB) instead of averaging scores of several snippets. Inspired by separable depth-wise convolution \cite{xception}, TXB encodes temporal dynamics in a separate channel-wise and temporal-wise 1D convolution manner for smaller model size and higher computation efficiency. Finally, TXB is convolution based rather than recurrent architecture, it is easily to be optimized via stochastic gradient decent (SGD) in an end-to-end manner.

We evaluate the proposed StNet framework over the newly released large scale action recognition dataset Kinetics \cite{kinetics}. Experiment results show that StNet outperforms several state-of-the-art 2D and 3D convolution based solutions, meanwhile our StNet attains better efficiency from the perspective of the number of FLOPs and higher effectiveness in terms of recognition accuracy than its 3D CNN counterparts. Besides, the learned representation of StNet is transferred to the UCF101 \cite{ucf101} dataset to verify its generalization capability.

\section{Related Work}
\label{sec:related}
In the literature, video-based action recognition solutions can be divided into two categories: action recognition with hand-crafted features and action recognition with deep ConvNet. To develop effective spatial-temporal representations, researchers have proposed many hand-crafted features such as HOG3D \cite{hog3d}, SIFT3D \cite{sift3d}, MBH \cite{mbh}. Currently, improved dense trajectory \cite{idt} is the state-of-the-art among the hand-crafted features. Despite its good performance, such hand-crafted feature is designed for local spatial-temporal description and is hard to capture semantic level concepts. 
Thanks to the big progress made by introducing deep convolution neural network, ConvNet based action recognition methods have achieved superior accuracy to conventional hand-crafted methods. As for utilizing CNN for video-based action recognition, there exist the following two research directions:

\textbf{Encoding CNN Features:} CNN is usually used to extract spatial features from video frames, and the extracted feature sequence is then modelled with recurrent neural networks or feature encoding methods.
In LRCN \cite{LRCN}, CNN features of video frames are fed into LSTM network for action classification. ShuttleNet \cite{shuttleNet} introduced biologically-inspired feedback connections to model long-term dependencies of spatial CNN descriptors. TLE \cite{tle} proposed temporal linear encoding that captures the interactions between video segments, and encoded the interactions into a compact representations. Similarly, VLAD \cite{netvlad,actionvlad} and AttentionClusters \cite{attentioncluster} have been proposed for local feature integration.

\textbf{ConvNet as Recognizer:} the first attempt to use deep convolution network for action recognition was made by Karpathy et.al, \cite{karpathy2014}. While strong results for action recognition have been achieved by \cite{karpathy2014}, two stream ConvNet \cite{two-stream} that merges the predicted scores from a RGB based spatial stream and an optical flow based temporal stream obtained performance improvements in a large margin. ST-ResNet \cite{two-stream-stresnet} introduced residual connections between the two streams of \cite{two-stream} and showed great advantage in results. To model the long-range temporal structure of videos, Temporal Segment Network (TSN) \cite{tsn} was proposed to enable efficient video-level supervision by sparse temporal sampling strategy and further boosted the performance of ConvNet based action recognizer.

\begin{figure*}[!ht]
\begin{center}
\includegraphics[width=0.8\textwidth]{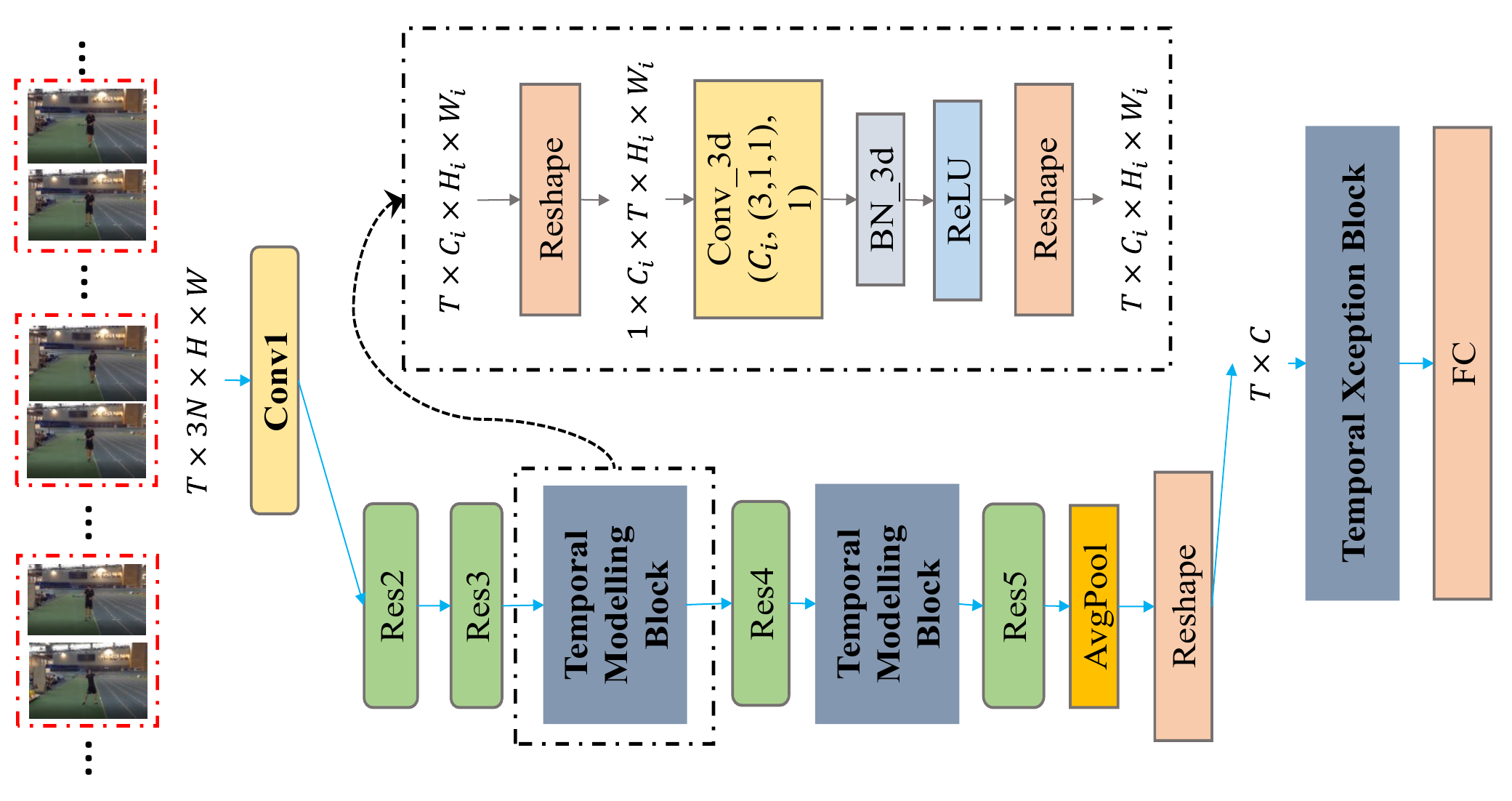}
\end{center}
   \caption{Illustration of constructing StNet based on ResNet \cite{resnet} backbone. The input to StNet is a $T\times 3N \times H \times W$ tensor. Local spatial-temporal patterns are modelled via 2D Convolution. 3D convolutions are inserted right after the Res3 and Res4 blocks for long term temporal dynamics modelling. The setting of 3D convolution (\# Output Channel, (temporal kernel size, height kernel size, width kernel size), \# groups) is ($C_i$, (3,1,1), 1). }
\label{fig:stnet}
\end{figure*}
Observing that 2D ConvNets cannot directly exploit the temporal patterns of actions, spatial-temporal modeling techniques were involved in a more explicit way. C3D \cite{c3d} applied 3D convolutional filters to the videos to learn spatial-temporal features. Compared to 2D ConvNets, C3D has more parameters and is much more difficult to obtain good convergence. To overcome this difficulty, T-ResNet \cite{T-Resnet} injects temporal shortcut connections between the layers of spatial ConvNets to get rid of 3D convolution. I3D \cite{i3d} simultaneously learns spatial-temporal representation from video by inflating conventional 2D ConvNet architecture into 3D ConvNet. P3D \cite{p3d} decouples a 3D convolution filter to a 2D spatial convolution filter followed by a 1D temporal convolution filter. Recently, there are many frameworks proposed to improve 3D convolution \cite{eco,ARN,s3d,r2+1d,nonlocal,mfnet}. Our work is different in that spatial-temporal relationship is progressively modeled via temporal convolution upon local spatial-temporal feature maps.


\section{Proposed Approach}
The proposed StNet can be constructed from the existing state-of-the-art 2D ConvNet frameworks, such as ResNet \cite{resnet}, InceptionResnet \cite{inceptionresnet} and so on. Taking ResNet as an example, Fig.\ref{fig:stnet} illustrates how we can build StNet from the existing 2D ConvNet. It is similar to build StNet from other 2D ConvNet frameworks such as InceptionResnetV2 \cite{inceptionresnet}, ResNeXt \cite{resnext} and SENet \cite{senet}. Therefore, we do not elaborate all such details here.

\begin{figure*}[!t]
\centering
\subfigure[Temporal Xception block configuration]{
\includegraphics[width=\columnwidth]{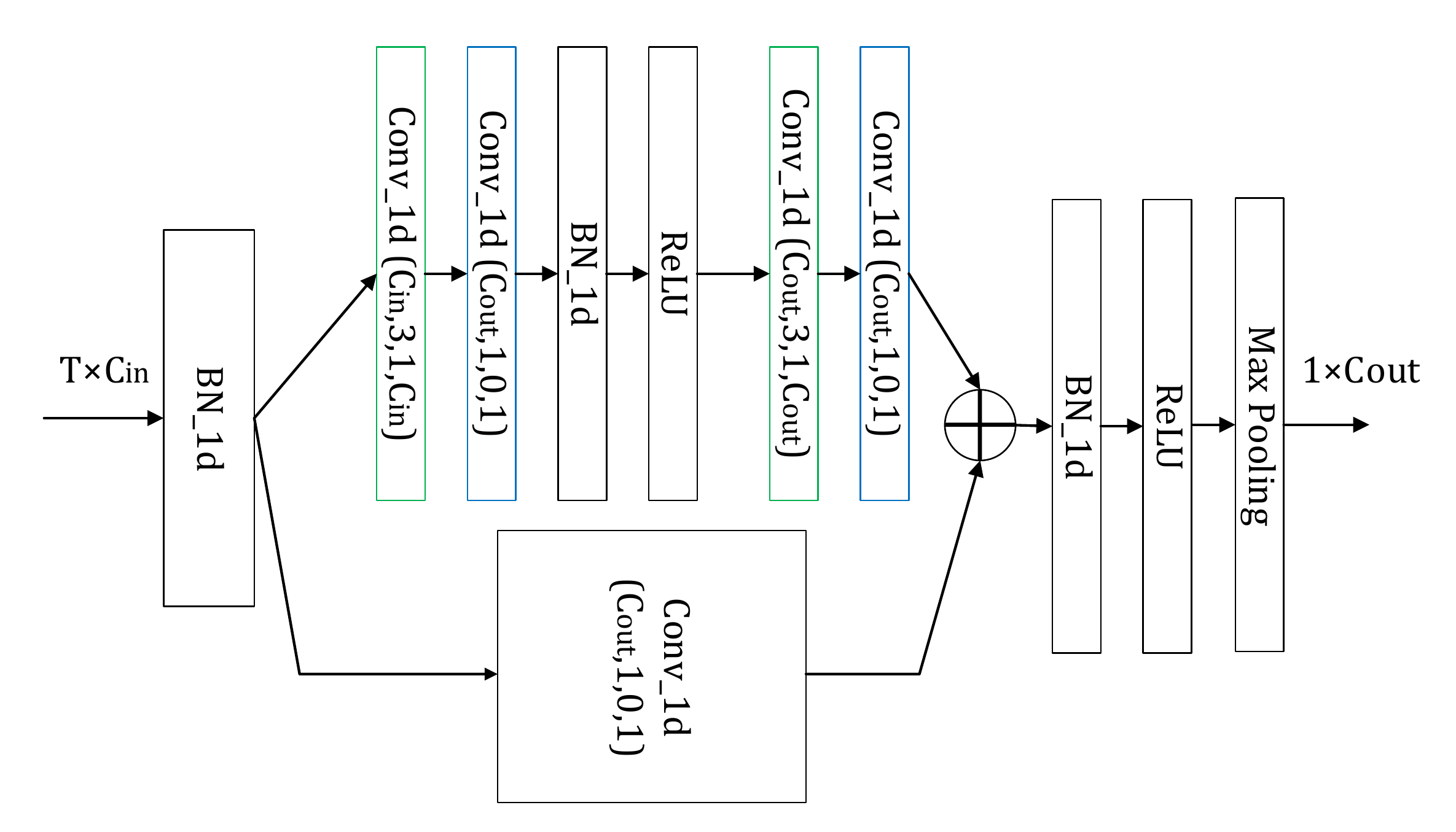}
\label{fig:txn}
}
\subfigure[Channel- and temporal-wise convolution]{
\includegraphics[width=\columnwidth]{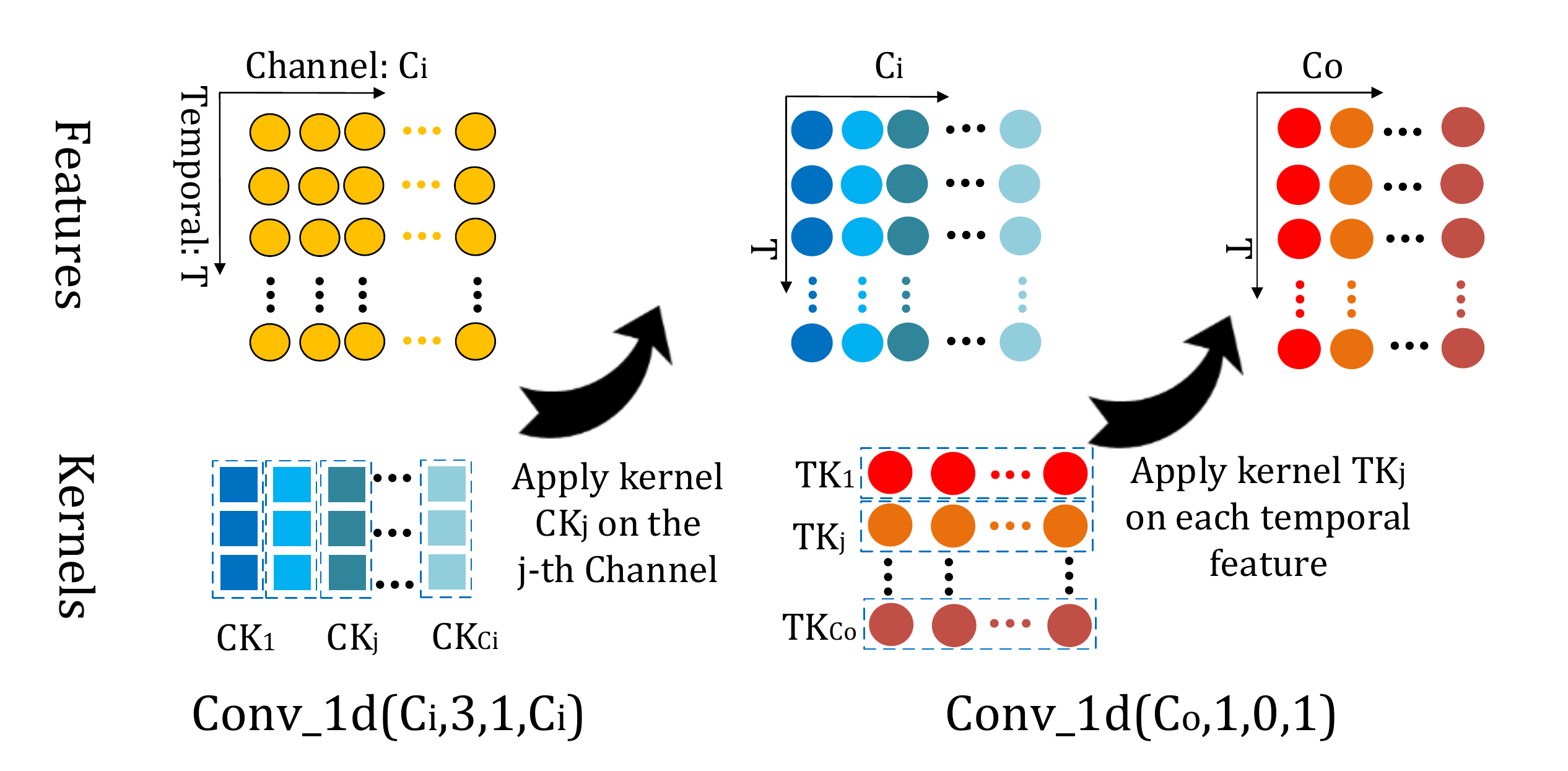}
\label{fig:separable}
}
   \caption{Temporal Xception block (TXB). The detailed configuration of our proposed temporal Xception block is shown in (a). The parameters in the bracket denotes (\#kernel, kernel size, padding, \#groups) configuration of 1D convolution. Blocks in green denote channel-wise 1D convolutions and blocks in blue denote temporal-wise 1D convolutions. (b) depicts the channel-wise and temporal-wise 1D convolution. Input to TXB is feature sequence of a video, which is denoted as a $T\times C_{in}$ tensor. Every kernel of channel-wise 1D convolution is applied along the temporal dimension within only one channel. Temporal-wise 1D convolution kernel convolves across all the channels along every temporal step.}
\label{fig:txn_approach}
\end{figure*}

\textbf{Super-Image}: Inspired by TSN \cite{tsn}, we choose to model long range temporal dynamics by sampling temporal snippets rather than inputting the whole video sequence. One of the differences from TSN is that we sample $T$ temporal segments each of which consists of $N$ consecutive RGB frames rather than a single frame. These $N$ frames are stacked in the channel dimension to form a \emph{super image}, so the input to the network is a tensor of size $T\times 3N \times H \times W$. Super-Image contains not only local spatial appearance information represented by individual frame but also local temporal dependency among these successive video frames. In order to jointly modeling the local spatial-temporal relationship therein and as well as to save model weights and computation costs, we leverage 2D convolution (whose input channel size is $3N$) on each of the $T$ super-images. Specifically, the local spatial-temporal correlation is modeled by 2D convolutional kernels inside the Conv1, Res2, and Res3 blocks of ResNet as shown in Fig.\ref{fig:stnet}. In our current setting, $N$ is set to 5. In the training phase, 2D convolution blocks can be initialized directly with weights from the ImageNet pre-trained backbone 2D convolution model except the first convolution layer. Weights of Conv1 can be initialized following what the authors have done in I3D \cite{i3d}.

\textbf{Temporal Modeling Block}: 2D convolution on the $T$ super-images generates $T$ local spatial-temporal feature maps. Building the global spatial-temporal representation of the sampled $T$ super-images is essential for understanding the whole video. Specifically, we choose to insert two temporal modeling blocks right after the Res3 and Res4 block. The temporal modeling blocks are designed to capture the long-range temporal dynamics inside a video sequence and they can be easily implemented by leveraging the architecture of Conv3d-BN3d-ReLU. Note that the existing 2D ConvNet framework is powerful enough for spatial modeling, so we set both spatial kernel size of a 3D convolution as 1 to save computation cost while the temporal kernel size is empirically set to be 3. Applying 2 temporal convolutions on the $T$ local spatial-temporal feature maps after Res3 and Res4 blocks introduces very limited extra computation cost but is effective to capture global spatial-temporal correlation progressively. In the temporal modeling blocks, weights of Conv3d layers are initially set to $1/(3\times C_i)$, where $C_i$ denotes input channel size, and biases are set to 0. BN3d is initialized to be an identity mapping.

\textbf{Temporal Xception Block}: Our temporal Xception block is designed for efficient temporal modeling among feature sequence and easy optimization in an end-to-end manner. We choose temporal convolution to capture temporal relations instead of recurrent architectures mainly for the end-to-end training purpose. Unlike ordinary 1D convolution which captures the channel-wise and temporal-wise information jointly, we decouple channel-wise and temporal-wise calculation for computational efficiency.

The temporal Xception architecture is shown in Fig.\ref{fig:txn}. The feature sequence is viewed as a $T\times C_{in}$ tensor, which is obtained by globally average pooling from the feature maps of $T$ super-images. Then, 1D batch normalization \cite{bn} along the channel dimension is applied to such an input to handle the well-known co-variance shift issue, the output signal is $V$,
\begin{equation}\label{eq:bn}
  v^{i} = \frac{u^{i}-m}{\sqrt{var}}*\alpha+\beta
\end{equation}
where $v^{i}$ and $u^{i}$ denote the $i^{th}$ row of the output and input signals, respectively; $\alpha$ and $\beta$ are trainable parameters, $m$ and $var$ are accumulated running mean and variance of input mini-batches. To model temporal relation, convolutions along the temporal dimension are applied to $V$. We decouple temporal convolution into separate channel-wise and temporal-wise 1D convolutions. Technically, for channel-wise 1D convolution, the temporal kernel size is set to 3, and the number of kernels and the group number are set to be the same with the input channel number. In this sense, every kernel convolves over temporal dimension within a single channel. For temporal-wise 1D convolution, we set both the kernel size and the group number to 1, so that temporal-wise convolution kernels operate across all elements along the channel dimension at each time step. Formally, channel-wise and temporal-wise convolution can be described with Eq.\ref{eq:c-conv} and Eq.\ref{eq:t-conv}, respectively,
\begin{equation}\label{eq:c-conv}
  y_{i,j} = \sum_{k=0}^{2}x_{k+i-1,j}*W_{j,k,0}^{(c)}+b_{j}^{(c)},
\end{equation}
\begin{equation}\label{eq:t-conv}
  y_{i,j} = \sum_{k=0}^{C_{i}}x_{i,k}*W_{j,0,k}^{(t)}+b_{j}^{(t)},
\end{equation}
where $x\in R^{T\times C_{i}}$ is the input $C_{i}$-$Dim$ feature sequence of length $T$, $y\in R^{T\times C_o}$ denotes output feature sequence and $y_{i,j}$ is the value of the $j^{th}$ channel of the $i^{th}$ feature, $*$ denotes multiplication. In Eq.\ref{eq:c-conv}, $W^{(c)}\in R^{C_o\times 3\times 1}$ denotes the channel-wise Conv kernel of (\#kernel, kernel size, \#groups) = ($C_o$,3,$C_i$). In Eq.\ref{eq:t-conv}, $W^{(t)}\in R^{C_{o}\times 1 \times C_i}$ denotes the temporal-wise Conv kernel of (\#kernel, kernel size, \#groups) = ($C_o$,1,1).
$b$ denotes bias. In this paper, $C_o$ is set to 1024. An intuitive illustration of separate channel- and temporal-wise convolution can be found in Fig.\ref{fig:separable}.

As shown in Fig.\ref{fig:txn}, similar to the bottleneck design of \cite{resnet}, the temporal Xception block has a long branch and a short branch. The short branch is a single 1D temporal-wise convolution whose kernel size and group size are both 1. Therefore, the short branch has a temporal receptive field of 1. Meanwhile, the long branch contains two channel-wise 1D convolution layers and thus has a temporal receptive filed of 5. The intuition is that, fusing branches with different temporal receptive field sizes is helpful for better temporal dynamics modeling. The output feature of the temporal Xception block is fed into a 1D max-pooling layer along the temporal dimension, and the pooled output is used as the spatial-temporal aggregated descriptor for classification.

\section{Experiments}
\subsection{Datasets and Evaluation Metric}
To evaluate the performance of our proposed StNet framework for large scale video-based action recognition, we perform extensive experiments on the recent large scale action recognition dataset named Kinetics \cite{kinetics}. The first version of this dataset (denoted as Kinetics400) has 400 human action classes, with more than 400 clips for each class. The validation set of Kinetics400 consists of about 20K video clips. The second version of Kinetics (denoted as Kinetics600) contains 600 action categories and there are about 400K trimmed video clips in its training set and 30K clips in the validation set. Due to unavailability of ground truth annotations for testing set,  the results on the Kinetics dataset in this paper are evaluated on its validation set.

To validate that the effectiveness of StNet could be transferred to other datasets, we conduct transfer learning experiments on the UCF101 \cite{ucf101}, which is much smaller than Kinetics. It contains 101 human action categories and 13,320 labeled video clips in total. The labeled video clips are divided into three training/testing splits for evaluation. In this paper, the evaluation metric for recognition effectiveness is average class accuracy, we also report total number of model parameters as well as FLOPs (total number of float-point multiplications executed in the inference phase) to depict model complexity.

\subsection{Ablation Study}
\begin{table}[t]
\centering
\begin{tabular}{ c | c | c}
  \hline\hline		
   Configuration & Top-1 & \# Params \\
  \hline
  TSN (Backbone) & 73.02 & - \\
  \hline
   w/o 1D BatchNorm  & 73.55 & - \\
   w/o C-Conv &74.14 &- \\
   w/o T-Conv &74.06 &- \\
   w/o Short-Branch  &74.33 &- \\
   w/o Long-Branch  &74.21 &- \\
   \hline
  Ordinary Temporal-Conv  & 74.28 &9.6M \\
  LSTM &73.21  &10.9M \\
  GRU &73.66  & 8.3M \\
  \hline
  proposed TXB &\textbf{74.62} & \textbf{4.6M} \\
  \hline
\end{tabular}
\caption{Ablation study of TXN on Kinetics400. C-Conv and T-Conv denote channel-wise and temporal-wise 1D Conv, respectively. Prec@1 and number of model parameters are reported in the table.}
\label{t:ablation1}
\end{table}

\subsubsection{Temporal Xception Block}
We conduct ablation experiments on our proposed temporal Xception block. To show the contribution of each component in TXB, we disable each of them one by one, and then train and test the models on the RGB feature sequence, which is extracted from the GlobalAvgPool layer of InceptionResnet-V2-TSN \cite{tsn} model trained on the Kinetics400 dataset. Besides, we also implemented an ordinary 2-layered temporal Conv model (denoted as Ordinary Temporal-Conv) and RNN-based models (LSTM \cite{lstm} and GRU \cite{gru}) for comparison. In Ordinary Temporal-Conv model, we replace the temporal Xception module by two 1D convolution layers, whose kernel size is 3 and output channel number is 1024, to temporally model the input feature sequence. In this experiment, the hidden units of LSTM and GRU is set to 1024. The final classification results are predicted by a fully connected layer with output size of 400. For each video, features of 25 frames are evenly sampled from the whole feature sequence to train and test all the models. The evaluation results and number of parameters of these models are reported in Table.\ref{t:ablation1}

From the top lines of Table.\ref{t:ablation1}, we can see that each component contributes to the proposed TXB framework. Batch normalization handles the co-variance shift issue and it brings 1.07\% absolute top-1 accuracy improvement for RGB stream. Separate channel-wise and temporal-wise convolution layers is helpful for modeling temporal relations, and recognition performance drops without either of them. The results also demonstrate that our design of long-branch plus short-branch is useful by mixing multiple temporal receptive field encodings. Comparing the results listed in the middle lines with that of our TXB, it is clear that TXB achieves the best top-1 accuracy among these models and the model is the smallest (with only 4.6 million parameters in total), especially, the gain over backbone TSN is up to 1.6 percent.

\subsubsection{Impact of Each Component in StNet}
In this section, a series of ablation studies are performed to understand the importance of each design choice of our proposed StNet. To this end, we train multiple variants of our model to show how the performance is improved with our proposed super-image, temporal modeling blocks and temporal Xception block, respectively. There are chances that some tricks would be effective when either shallow backbone networks are used or evaluating on small datasets, therefore we choose to carry out experiments on the very large Kinetics600 dataset and the backbone we used is the very deep and powerful InceptionResnet-V2 \cite{inceptionresnet}. Super-Image (SI), temporal modeling blocks (TM), and temporal Xception block (TXB) are enabled one after another to generate 4 network variants and in this experiment, $T$ is set to 7 and $N$ to 5. The video frames are scaled such that their short-size is 331 and a random and the central $299\times299$ patch is cropped from each of the $T$ frames in the training phase and testing phase, respectively.

\begin{table}[!t]
\centering
\begin{tabular}{ c | c | c |c }
  \hline\hline		
  \multicolumn{3}{c|} {Configurations} & \multirow{2}{*} {Top-1}\\	
  \cline{1-3}
  Super-Image & TM Blocks & TXB & \\
  \hline
  $\times$ ~(N=1) &$\times$ &$\times$  & 72.2 \\
  $\surd$ ~(N=5) & $\times$ & $\times$ &74.2 \\
  $\surd$ ~(N=5) & $\surd$ &$\times$   &76.0  \\
  $\surd$ ~(N=5) & $\surd$ & $\surd$ &76.3 \\
  \hline
\end{tabular}
\caption{Results evaluated on Kinetics600 validation set with different network configurations and $T=7$ .}
\label{t:ablation2}
\end{table}

\begin{table*}[t]
 \centering
 \begin{tabular}{ c | c | c |c |c |c |c }
 \hline\hline
 Framework          & Backbone           & Input $\times$ \# Clips                            & Dataset & Prec@1   & \# Params & FLOPs \\
 \hline
 \multirow{2}{*}{C2D \cite{nonlocal}} & ResNet50              & [32$\times$3$\times$256$\times$256]$\times$1 & \multirow{15}{*}{K400}  &\color{red}{$62.42$} & \multirow{2}{*}{24.27M} & \color{green}{26.29G}  \\
                    & ResNet50              & [32$\times$3$\times$256$\times$256]$\times$10 &         &$69.90$ & &262.9G  \\
  \cline{1-3}\cline{5-7}
  \multirow{3}{*}{C3D \cite{nonlocal}} & ResNet50            & [32$\times$3$\times$256$\times$256]$\times$1 &         &\color{red}{$64.65$} &\multirow{2}{*}{35M} &164.84G \\
                       & ResNet50            & [32$\times$3$\times$256$\times$256]$\times$10 &         &\color{green}{$71.86$} & &\color{red}{1648.4G} \\
   \cline{1-3}\cline{5-7}
  I3D (\citeauthor{i3d})                     & BN-Inception     & [All$\times$3$\times$256$\times$256]$\times$1 &         &$70.24$ &\color{green}{12.7M} &\color{red}{544.44G} \\
  \cline{1-3}\cline{5-7}
  S3D(\citeauthor{s3d}) & BN-Inception & [All$\times$3$\times$224$\times$224]$\times$1 & &\color{green}{72.20} &\color{green}{8.8M} &\color{red}{518.6G} \\
  \cline{1-3}\cline{5-7}
  \multirow{2}{*}{MF-Net(\citeauthor{mfnet})} & \multirow{2}{*}{-} & [16$\times$3$\times$224$\times$224]$\times$1 & &\color{red}{65.00} &\multirow{2}{*}{\color{green}{8.0M}} &\color{green}{11.1G} \\
                                              &                    & [16$\times$3$\times$224$\times$224]$\times$50 & &\color{green}{72.80} &                         &\color{red}{555G} \\
  \cline{1-3}\cline{5-7}
  R(2+1)D-RGB(\citeauthor{r2+1d}) & ResNet34 & [32$\times$3$\times$112$\times$112]$\times$10 & &\color{green}{72.00} &63.8M &\color{red}{1524G} \\
  \cline{1-3}\cline{5-7}
  \cline{1-3}\cline{5-7}
  \multirow{2}{*}{Nonlocal-I3d(\citeauthor{nonlocal})} & \multirow{2}{*}{ResNet50} & [128$\times$3$\times$224$\times$224]$\times$1 & &\color{red}{67.30} &\multirow{2}{*}{35.33M} &145.7G \\
                                             &  & [128$\times$3$\times$224$\times$224]$\times$30 & &\color{green}{76.50} & &\color{red}{4371G} \\
  \cline{1-3}\cline{5-7}

  \multirow{2}{*}{\textbf{StNet (Ours)}} &ResNet50           & [25$\times$15$\times$256$\times$256]$\times$1 &      &\textbf{69.85} &\textbf{33.16M} &\textbf{189.29G} \\
                         &ResNet101           & [25$\times$15$\times$256$\times$256]$\times$1 &     &\textbf{71.38} &\textbf{52.15M} &\textbf{310.50G} \\

  \hline\hline
  \multirow{2}{*}{TSN \cite{tsn}} &IRv2      & [25$\times$3$\times$331$\times$331]$\times$1 &  \multirow{7}{*}{K600}   &76.22 &55.23M &410.85G \\
                       &SE-ResNeXt152           & [25$\times$3$\times$256$\times$256]$\times$1 &    &76.16 &\color{red}{142.94M} &\color{red}{875.21G} \\
  \cline{1-3}\cline{5-7}
  I3D (\citeauthor{carreira2018short})  & BN-Inception & [All$\times$3$\times$256$\times$256]$\times$1 &         &\color{red}{$71.9$} &\color{green}{12.90M} &544.45G \\
  \cline{1-3}\cline{5-7}
  \multirow{2}{*}{P3D \cite{yao2018yh}} &ResNet152           & [32$\times$3$\times$299$\times$299]$\times$1 &        &\color{red}{71.31} &66.90M &\color{green}{132.38G} \\
                       & U                    & [128$\times$3$\times$U$\times$U]$\times$U   &       &\color{green}{$77.94$} & U  & - \\
  \cline{1-3}\cline{5-7}
  \multirow{2}{*}{\textbf{StNet (Ours)}} &SE-ResNeXt101           & [25$\times$15$\times$256$\times$256]$\times$1 &    &76.04 &79.13M &453.95G \\
                         &IRv2      & [25$\times$15$\times$331$\times$331]$\times$1 &  &\color{green}{\textbf{78.99}} &\textbf{72.13M} &\textbf{439.57G} \\

 \hline\hline
 \end{tabular}
 \caption{Comparison of StNet and several state-of-the-art 2D/3D convolution based solutions. The results are reported on validation set of Kinetics400 and Kinetics600, with RGB modality only. We investigate both Prec@1 and model efficiency w.r.t. total number of model parameters and FLOPs needed in inference. Here, ``IRv2'' denotes InceptionResNet-V2, ``K400'' is short for Kinetics400 and so is the case for ``K600''. U denotes unknown. ``All'' means using all frames in a video.}
 \label{t:compare}
\end{table*}
Experiment results are reported in Table.\ref{t:ablation2}. When the three components are all disabled, the model degrades to be TSN \cite{tsn}, which achieves top-1 precision of 72.2\% on Kinetics600 validation set. By enabling super-image, the recognition performance is improved by 2.0\% and the gain comes from introducing local spatial-temporal modelling. When the two temporal modelling blocks are inserted, the Prec@1 is further boosted to 76.0\%, it evidences the fact that modeling global spatial-temporal interactions among the feature maps of super-images is necessary for performance improvement, because it can well represent high-level video features. The final performance is 76.3\% when all the components are integrated and this shows that using TXB to capture long-term temporal dynamics is still a plus even if local and global spatial-temporal relationship is modeled by enabling super-images and temporal modeling blocks.

\subsection{Comparison with Other Methods}
We evaluate the proposed framework against the recent state-of-the-art 2D/3D convolution based solutions. Extensive experiments are conducted on the datasets of Kinetics400 and Kinetics600 to make comparisons among these models in terms of their effectiveness (i.e., top-1 accuracy) and efficiency (reflected by total number of model parameters and FLOPs needed in the inference phase). To make thorough comparison, we evaluated different methods with several relatively small backbone networks and a few very deep backbones on Kinetics400 and Kinetics600, respectively. Results are summarized in Table.\ref{t:compare}. Numbers in green mean that the results are pleasing and numbers in red represent unsatisfying ones. From the evaluation results, we can draw the following conclusions:

\textbf{StNet outperforms 2D-Conv based solution:} (1) C2D-ResNet50 is cheap in FLOPs, but its top-1 recognition precision is very poor (62.42\%) if only 1 clip is tested. When 10 clips are tested, the performance is boosted to 69.9\% at the cost of 262.9G FLOPs. StNet-ResNet50 achieves Prec@1 of 69.85\% and only 189.29G FLOPs are needed. (2) When large backbone models are used, StNet still outperforms 2D-Conv based solution, this can be concluded from the fact that StNet-IRv2 significantly boosts the performance of TSN-IRv2 (which is 76.22\%) to 78.99\% while the total number of FLOPs is slightly increased from 410.85G to 439.57G. Besides, StNet-SE-ResNeXt101 performs comparable with TSN-SE-ResNeXt152, but the model size and FLOPs are significantly saved (from 875.21G to 453.95G).

\begin{figure*}[!ht]
\centering
\subfigure[Activation maps of TSN]{
\includegraphics[width=\columnwidth]{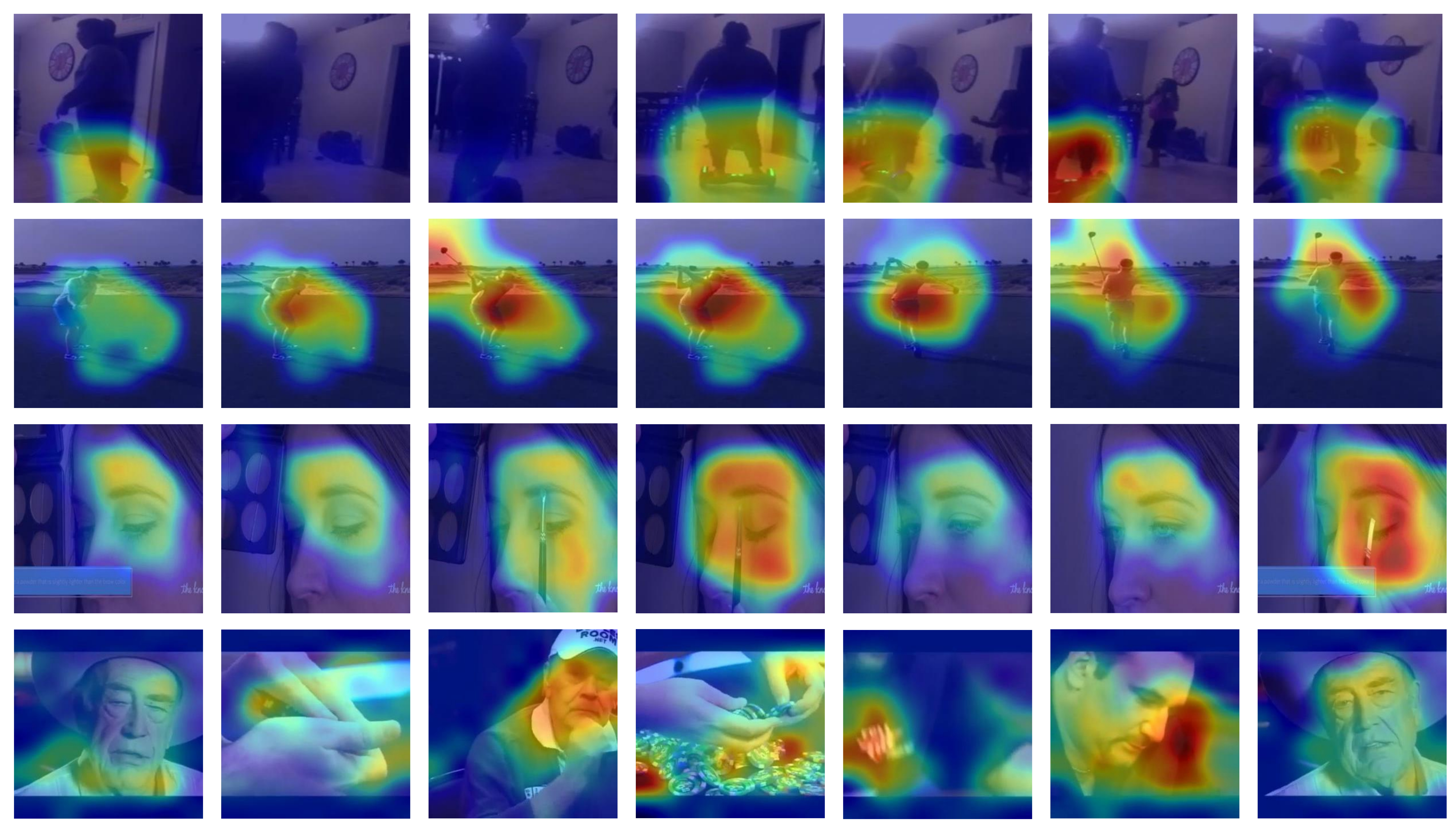}
\label{fig:visualize_tsn}
}
\subfigure[Activation maps of StNet]{
\includegraphics[width=\columnwidth]{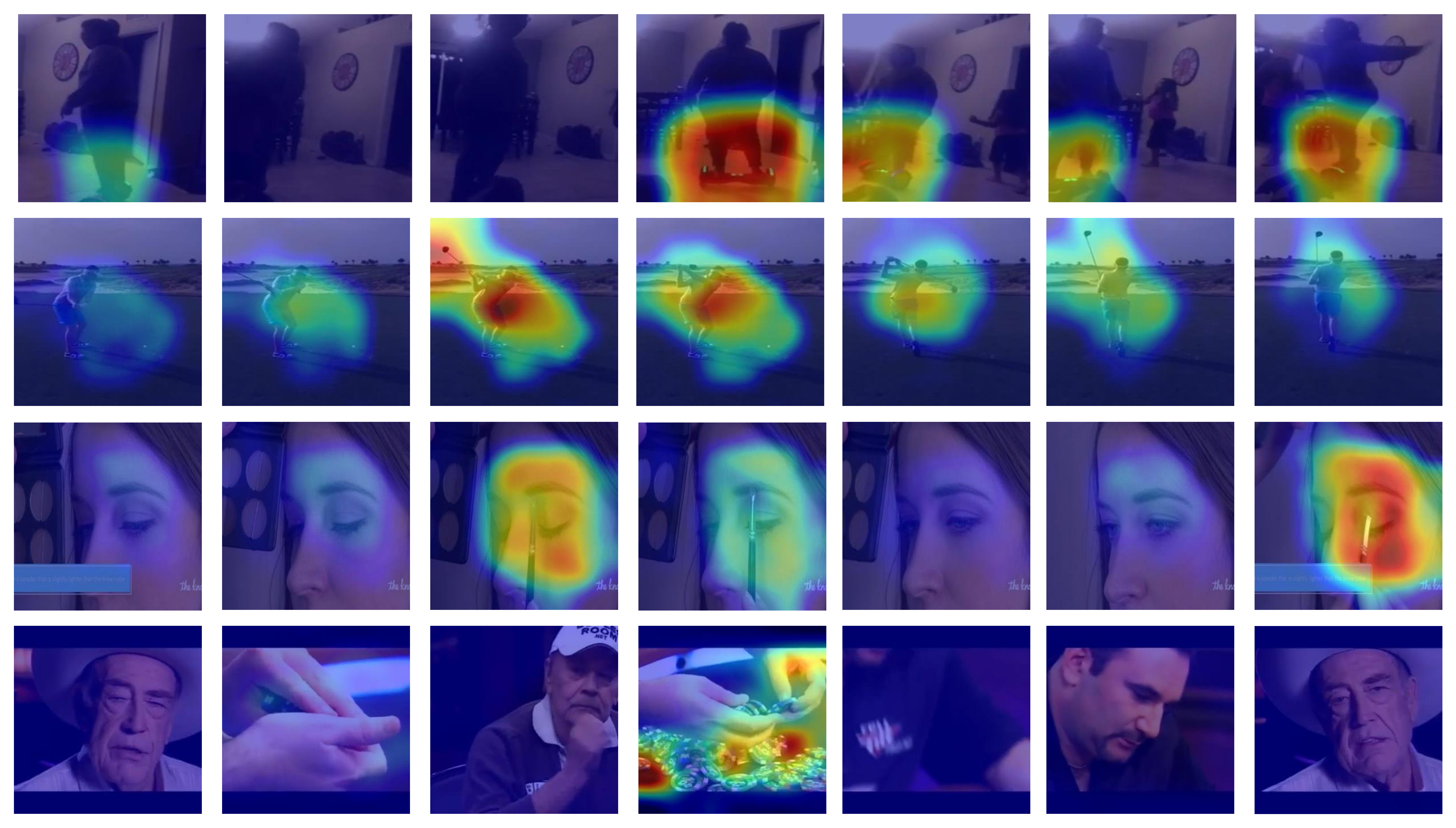}
\label{fig:visualize_xception}
}
   \caption{Visualizing action-specific activation maps with the CAM \cite{cam} approach. For illustration, activation maps of video snippets of four action classes, i.e., hoverboarding, golf driving ,filling eyebrows and playing poker, are shown from top to the bottom. It is clear that StNet can well capture temporal dynamics in video and focuses on the spatial-temporal regions which really corresponds to the action class.}
\label{fig:visualize}
\end{figure*}

\textbf{StNet obtains a better trade off than 3D-Conv:} (1) Though model size and recognition performance of I3D is very plausible, it is very costly. C3D-ResNet50 achieves Prec@1 of 64.65\% with single clip test. With comparable FLOPs, StNet-ResNet50 achieves 69.85\% which is much better than C3D-ResNet50. Ten clip test significantly improves performance to 71.86\%, however, the FLOPs is as huge as 1648.4G. StNet-ResNet101 achieves 71.38\% with over 5x FLOPs reduction. (2) Compared with P3D-ResNet152, StNet-IRv2 outperforms by a large margin (78.99\% v.s. 71.31) with acceptable FLOPs increase (from 132.38G to 439.57G). Besides, Single clip test performance of StNet-IRv2 still outperforms P3D by 1.05\%, which is used for the Kinetics600 challenge with 128 frames input and no further details about its backbone, input size as well as number of testing clips.
(3) Compared with S3D \cite{s3d}, R(2+1)D\cite{r2+1d}, MF-Net \cite{mfnet} and Nonlocal-I3d \cite{nonlocal}, the proposed StNet can still strike good performance-FLOPs trade-off.

\subsection{Transfer Learning on UCF101}

We transfer RGB models of StNet pre-trained on Kinetics to the much smaller dataset of UCF101 \cite{ucf101} to show that the learned representation can be well generalized to other dataset. The results included in Table \ref{t:transfer} are the mean class accuracy from three training/testing splits. It is clear that our Kinetics pre-trained StNet models with ResNet50, ResNet101 and the large InceptionResNet-V2 backbone demonstrate very powerful transfer learning capability, and mean class accuracy is up to 93.5\%, 94.3\% and 95.7\%, respectively. Specifically, the transferred StNet-IRv2 RGB model achieves the state-of-the-art performance while its FLOPs is 123G.
\begin{table}[h]
 \centering
 \begin{tabular}{ c | c | c |c }
 \hline\hline
 Model & Pre-Train & FLOPs & Accuracy \\
 \hline
 C3D+Res18                                        &K400 &- &89.8 \\
 \hline
 I3D+BNInception                 &K400  & \color{red}{544G} &{95.6} \\
 \hline
TSN+BNInception &K400 &- &91.1 \\
   TSN+IRv2 (T=25)        &K400 &411G &92.7 \\
 \hline
 StNet-Res50 (T=7) & K400 &\textbf{\color{green}{53G}} &\textbf{93.5} \\
 StNet+Res101 (T=7)  &K400 &\textbf{87G} &\textbf{94.3} \\
 StNet+IRv2 (T=7)       &K600 &\textbf{123G} &\textbf{95.7} \\
 \hline\hline
 \end{tabular}
 \caption{Mean class accuracy achieved by different model transfer learning experiments. RGB frames of UCF101 are used for training and testing. The mean class accuracy averaged over the three splits of UCF101 is reported.}
 \label{t:transfer}
\end{table}

\subsection{Visualization in StNet}
To help us better understand how StNet learns discriminative spatial-temporal descriptors for action recognition, we visualize the class-specific activation maps of our model with the CAM \cite{cam} approach. In this experiment, we set $T$ to 7 and the 7 snippets are evenly sampled from video sequences in the Kinetics600 validation set to obtain their class-specific activation map. As a comparison, we also visualize the activation maps of TSN model, which exploits local spatial information and then fuses $T$ snippets by averaging classification score rather than jointly modeling local and global spatial-temporal information. As an illustration, Fig. \ref{fig:visualize} lists activation maps of four action classes (hoverboarding, golf driving, filling eyebrows and playing poker) of both models.

These maps shows that, compared to TSN which fails to jointly model local and global spatial-temporal dynamics in videos, our StNet can capture the temporal interactions inside video frames. It focuses on the spatial-temporal regions which are closely related to the groundtruth action. For example, it pays more attention to faces with eyebrow pencil in the nearby while regions with only faces are not so activated. Particularly, in the ``play poker'' example, StNet is significantly activated only by the hands and casino tokens. Nevertheless, TSN is activated by many regions of faces.

\section{Conclusion}
In this paper, we have proposed the StNet for joint local and global spatial-temporal modeling to tackle the action recognition problem. Local spatial-temporal information is modeled by applying 2D convolutions on sampled super-images, while the global temporal interactions are encoded by temporal convolutions on local spatial-temporal feature maps of super-images. Besides, we propose temporal Xception block to further modeling temporal dynamics. Such design choices make our model relatively lightweight and computation efficient in the training and inference phases. So it allows leveraging powerful 2D CNN to better explore large scale dataset. Extensive experiments on large scale action recognition benchmark Kinetics have verified the effectiveness of StNet. In addition, StNet trained on Kinetics exhibits pretty good transfer learning ability on the UCF101 dataset.

\bibliographystyle{./aaai}

\small

\begin{thebibliography}{}

\bibitem[\protect\citeauthoryear{Arandjelovic \bgroup et al\mbox.\egroup
  }{2016}]{netvlad}
Arandjelovic, R.; Gronat, P.; Torii, A.; Pajdla, T.; and Sivic, J.
\newblock 2016.
\newblock Netvlad: Cnn architecture for weakly supervised place recognition.
\newblock In {\em Proceedings of the IEEE Conference on CVPR},  5297--5307.

\bibitem[\protect\citeauthoryear{Carreira and Zisserman}{2017}]{i3d}
Carreira, J., and Zisserman, A.
\newblock 2017.
\newblock Quo vadis, action recognition? a new model and the kinetics dataset.
\newblock In {\em The IEEE Conference on CVPR}.

\bibitem[\protect\citeauthoryear{Carreira \bgroup et al\mbox.\egroup
  }{2018}]{carreira2018short}
Carreira, J.; Noland, E.; Banki-Horvath, A.; Hillier, C.; and Zisserman, A.
\newblock 2018.
\newblock A short note about kinetics-600.
\newblock {\em arXiv preprint arXiv:1808.01340}.

\bibitem[\protect\citeauthoryear{Chen \bgroup et al\mbox.\egroup
  }{2018}]{mfnet}
Chen, Y.; Kalantidis, Y.; Li, J.; Yan, S.; and Feng, J.
\newblock 2018.
\newblock Multi-fiber networks for video recognition.
\newblock In {\em The European Conference on Computer Vision (ECCV)}.

\bibitem[\protect\citeauthoryear{Cho \bgroup et al\mbox.\egroup }{2014}]{gru}
Cho, K.; Van~Merri{\"e}nboer, B.; Bahdanau, D.; and Bengio, Y.
\newblock 2014.
\newblock On the properties of neural machine translation: Encoder-decoder
  approaches.
\newblock {\em arXiv preprint arXiv:1409.1259}.

\bibitem[\protect\citeauthoryear{Chollet}{2017}]{xception}
Chollet, F.
\newblock 2017.
\newblock Xception: Deep learning with depthwise separable convolutions.
\newblock In {\em The IEEE Conference on CVPR}.

\bibitem[\protect\citeauthoryear{Dalal, Triggs, and Schmid}{2006}]{mbh}
Dalal, N.; Triggs, B.; and Schmid, C.
\newblock 2006.
\newblock Human detection using oriented histograms of flow and appearance.
\newblock In {\em European conference on computer vision},  428--441.
\newblock Springer.

\bibitem[\protect\citeauthoryear{Diba, Sharma, and Van~Gool}{2017}]{tle}
Diba, A.; Sharma, V.; and Van~Gool, L.
\newblock 2017.
\newblock Deep temporal linear encoding networks.
\newblock In {\em The IEEE Conference on CVPR}.

\bibitem[\protect\citeauthoryear{Donahue \bgroup et al\mbox.\egroup
  }{2015}]{LRCN}
Donahue, J.; Anne~Hendricks, L.; Guadarrama, S.; Rohrbach, M.; Venugopalan, S.;
  Saenko, K.; and Darrell, T.
\newblock 2015.
\newblock Long-term recurrent convolutional networks for visual recognition and
  description.
\newblock In {\em Proceedings of the IEEE conference on CVPR},  2625--2634.

\bibitem[\protect\citeauthoryear{Feichtenhofer, Pinz, and
  Wildes}{2016}]{two-stream-stresnet}
Feichtenhofer, C.; Pinz, A.; and Wildes, R.
\newblock 2016.
\newblock Spatiotemporal residual networks for video action recognition.
\newblock In {\em Advances in Neural Information Processing Systems},
  3468--3476.

\bibitem[\protect\citeauthoryear{Feichtenhofer, Pinz, and
  Wildes}{2017}]{T-Resnet}
Feichtenhofer, C.; Pinz, A.; and Wildes, R.~P.
\newblock 2017.
\newblock Temporal residual networks for dynamic scene recognition.
\newblock In {\em Proceedings of the IEEE Conference on CVPR},  4728--4737.

\bibitem[\protect\citeauthoryear{Fernando \bgroup et al\mbox.\egroup
  }{2015}]{videodarwin}
Fernando, B.; Gavves, E.; Oramas, J.~M.; Ghodrati, A.; and Tuytelaars, T.
\newblock 2015.
\newblock Modeling video evolution for action recognition.
\newblock In {\em Proceedings of the IEEE Conference on CVPR},  5378--5387.

\bibitem[\protect\citeauthoryear{Girdhar \bgroup et al\mbox.\egroup
  }{2017}]{actionvlad}
Girdhar, R.; Ramanan, D.; Gupta, A.; Sivic, J.; and Russell, B.
\newblock 2017.
\newblock Actionvlad: Learning spatio-temporal aggregation for action
  classification.
\newblock In {\em The IEEE Conference on CVPR}.

\bibitem[\protect\citeauthoryear{He \bgroup et al\mbox.\egroup }{2016}]{resnet}
He, K.; Zhang, X.; Ren, S.; and Sun, J.
\newblock 2016.
\newblock Deep residual learning for image recognition.
\newblock In {\em Proceedings of the IEEE conference on computer vision and
  pattern recognition},  770--778.

\bibitem[\protect\citeauthoryear{Hochreiter and Schmidhuber}{1997}]{lstm}
Hochreiter, S., and Schmidhuber, J.
\newblock 1997.
\newblock Long short-term memory.
\newblock {\em Neural computation} 9(8):1735--1780.

\bibitem[\protect\citeauthoryear{Hu, Shen, and Sun}{2017}]{senet}
Hu, J.; Shen, L.; and Sun, G.
\newblock 2017.
\newblock Squeeze-and-excitation networks.
\newblock {\em arXiv preprint arXiv:1709.01507}.

\bibitem[\protect\citeauthoryear{Ioffe and Szegedy}{2015}]{bn}
Ioffe, S., and Szegedy, C.
\newblock 2015.
\newblock Batch normalization: Accelerating deep network training by reducing
  internal covariate shift.
\newblock In {\em International Conference on Machine Learning},  448--456.

\bibitem[\protect\citeauthoryear{Karpathy \bgroup et al\mbox.\egroup
  }{2014}]{karpathy2014}
Karpathy, A.; Toderici, G.; Shetty, S.; Leung, T.; Sukthankar, R.; and Fei-Fei,
  L.
\newblock 2014.
\newblock Large-scale video classification with convolutional neural networks.
\newblock In {\em CVPR},  1725--1732.

\bibitem[\protect\citeauthoryear{Kay \bgroup et al\mbox.\egroup
  }{2017}]{kinetics}
Kay, W.; Carreira, J.; Simonyan, K.; Zhang, B.; Hillier, C.; Vijayanarasimhan,
  S.; Viola, F.; Green, T.; Back, T.; Natsev, P.; et~al.
\newblock 2017.
\newblock The kinetics human action video dataset.
\newblock {\em arXiv preprint arXiv:1705.06950}.

\bibitem[\protect\citeauthoryear{Klaser, Marsza{\l}ek, and
  Schmid}{2008}]{hog3d}
Klaser, A.; Marsza{\l}ek, M.; and Schmid, C.
\newblock 2008.
\newblock A spatio-temporal descriptor based on 3d-gradients.
\newblock In {\em BMVC 2008-19th British Machine Vision Conference},  275--1.
\newblock British Machine Vision Association.

\bibitem[\protect\citeauthoryear{Long \bgroup et al\mbox.\egroup
  }{2017}]{attentioncluster}
Long, X.; Gan, C.; de~Melo, G.; Wu, J.; Liu, X.; and Wen, S.
\newblock 2017.
\newblock Attention clusters: Purely attention based local feature integration
  for video classification.
\newblock {\em arXiv preprint arXiv:1711.09550}.

\bibitem[\protect\citeauthoryear{Qiu, Yao, and Mei}{2017}]{p3d}
Qiu, Z.; Yao, T.; and Mei, T.
\newblock 2017.
\newblock Learning spatio-temporal representation with pseudo-3d residual
  networks.
\newblock In {\em ICCV}.

\bibitem[\protect\citeauthoryear{Scovanner, Ali, and Shah}{2007}]{sift3d}
Scovanner, P.; Ali, S.; and Shah, M.
\newblock 2007.
\newblock A 3-dimensional sift descriptor and its application to action
  recognition.
\newblock In {\em Proceedings of the 15th ACM international conference on
  Multimedia},  357--360.
\newblock ACM.

\bibitem[\protect\citeauthoryear{Shi \bgroup et al\mbox.\egroup
  }{2017}]{shuttleNet}
Shi, Y.; Tian, Y.; Wang, Y.; Zeng, W.; and Huang, T.
\newblock 2017.
\newblock Learning long-term dependencies for action recognition with a
  biologically-inspired deep network.
\newblock In {\em ICCV}.

\bibitem[\protect\citeauthoryear{Simonyan and Zisserman}{2014}]{two-stream}
Simonyan, K., and Zisserman, A.
\newblock 2014.
\newblock Two-stream convolutional networks for action recognition in videos.
\newblock In {\em Advances in neural information processing systems},
  568--576.

\bibitem[\protect\citeauthoryear{Soomro, Zamir, and Shah}{2012}]{ucf101}
Soomro, K.; Zamir, A.~R.; and Shah, M.
\newblock 2012.
\newblock Ucf101: A dataset of 101 human actions classes from videos in the
  wild.
\newblock {\em arXiv preprint arXiv:1212.0402}.

\bibitem[\protect\citeauthoryear{Szegedy \bgroup et al\mbox.\egroup
  }{2017}]{inceptionresnet}
Szegedy, C.; Ioffe, S.; Vanhoucke, V.; and Alemi, A.~A.
\newblock 2017.
\newblock Inception-v4, inception-resnet and the impact of residual connections
  on learning.
\newblock In {\em AAAI},  4278--4284.

\bibitem[\protect\citeauthoryear{Tran \bgroup et al\mbox.\egroup }{2015}]{c3d}
Tran, D.; Bourdev, L.; Fergus, R.; Torresani, L.; and Paluri, M.
\newblock 2015.
\newblock Learning spatiotemporal features with 3d convolutional networks.
\newblock In {\em Proceedings of the IEEE ICCV},  4489--4497.

\bibitem[\protect\citeauthoryear{Tran \bgroup et al\mbox.\egroup
  }{2018}]{r2+1d}
Tran, D.; Wang, H.; Torresani, L.; Ray, J.; LeCun, Y.; and Paluri, M.
\newblock 2018.
\newblock A closer look at spatiotemporal convolutions for action recognition.
\newblock In {\em The IEEE Conference on CVPR}.

\bibitem[\protect\citeauthoryear{Wang and Schmid}{2013}]{idt}
Wang, H., and Schmid, C.
\newblock 2013.
\newblock Action recognition with improved trajectories.
\newblock In {\em Proceedings of the IEEE ICCV},  3551--3558.

\bibitem[\protect\citeauthoryear{Wang \bgroup et al\mbox.\egroup }{2016}]{tsn}
Wang, L.; Xiong, Y.; Wang, Z.; Qiao, Y.; Lin, D.; Tang, X.; and Van~Gool, L.
\newblock 2016.
\newblock Temporal segment networks: Towards good practices for deep action
  recognition.
\newblock In {\em European Conference on Computer Vision},  20--36.
\newblock Springer.

\bibitem[\protect\citeauthoryear{Wang \bgroup et al\mbox.\egroup }{2017a}]{ARN}
Wang, L.; Li, W.; Li, W.; and Van~Gool, L.
\newblock 2017a.
\newblock Appearance-and-relation networks for video classification.
\newblock {\em arXiv preprint arXiv:1711.09125}.

\bibitem[\protect\citeauthoryear{Wang \bgroup et al\mbox.\egroup
  }{2017b}]{nonlocal}
Wang, X.; Girshick, R.; Gupta, A.; and He, K.
\newblock 2017b.
\newblock Non-local neural networks.
\newblock {\em arXiv preprint arXiv:1711.07971}.

\bibitem[\protect\citeauthoryear{Xie \bgroup et al\mbox.\egroup
  }{2017}]{resnext}
Xie, S.; Girshick, R.; Doll{\'a}r, P.; Tu, Z.; and He, K.
\newblock 2017.
\newblock Aggregated residual transformations for deep neural networks.
\newblock In {\em Computer Vision and Pattern Recognition (CVPR), 2017 IEEE
  Conference on},  5987--5995.
\newblock IEEE.

\bibitem[\protect\citeauthoryear{Xie \bgroup et al\mbox.\egroup }{2018}]{s3d}
Xie, S.; Sun, C.; Huang, J.; Tu, Z.; and Murphy, K.
\newblock 2018.
\newblock Rethinking spatiotemporal feature learning: Speed-accuracy trade-offs
  in video classification.
\newblock In {\em The European Conference on Computer Vision (ECCV)}.

\bibitem[\protect\citeauthoryear{Yao and Li}{2018}]{yao2018yh}
Yao, T., and Li, X.
\newblock 2018.
\newblock Yh technologies at activitynet challenge 2018.
\newblock {\em arXiv preprint arXiv:1807.00686}.

\bibitem[\protect\citeauthoryear{Yue-Hei~Ng \bgroup et al\mbox.\egroup
  }{2015}]{yue2015beyond}
Yue-Hei~Ng, J.; Hausknecht, M.; Vijayanarasimhan, S.; Vinyals, O.; Monga, R.;
  and Toderici, G.
\newblock 2015.
\newblock Beyond short snippets: Deep networks for video classification.
\newblock In {\em Proceedings of the IEEE conference on CVPR},  4694--4702.

\bibitem[\protect\citeauthoryear{Zhou \bgroup et al\mbox.\egroup }{2016}]{cam}
Zhou, B.; Khosla, A.; Lapedriza, A.; Oliva, A.; and Torralba, A.
\newblock 2016.
\newblock Learning deep features for discriminative localization.
\newblock In {\em CVPR},  2921--2929.

\bibitem[\protect\citeauthoryear{Zolfaghari, Singh, and Brox}{2018}]{eco}
Zolfaghari, M.; Singh, K.; and Brox, T.
\newblock 2018.
\newblock Eco: Efficient convolutional network for online video understanding.
\newblock {\em arXiv preprint arXiv:1804.09066}.

\end{thebibliography}

\end{document}